\documentclass[12pt]{spieman}  
\usepackage{amsmath,amsfonts,amssymb}
\usepackage{graphicx}
\usepackage{setspace}
\usepackage{amsmath}
\usepackage{amssymb}
\usepackage{multirow}
\usepackage{float}
\usepackage{subcaption}
\usepackage{epstopdf}
\usepackage{color,soul}

\title{Discovery Radiomics via Evolutionary Deep Radiomic Sequencer Discovery for Pathologically-Proven Lung Cancer Detection}

\author[1*]{Mohammad Javad Shafiee}
\author[1]{Audrey G. Chung}
\author[2]{Farzad Khalvati}
\author[2]{Masoom A. Haider}
\author[1]{Alexander Wong}
\affil[1]{Vision and Image Processing Research Group, University of Waterloo, Waterloo, Canada}
\affil[2]{Department of Medical Imaging, University of Toronto and Sunnybrook Research Institute, Toronto, Canada}

\begin{document}
\maketitle

\begin{abstract}
\label{Abstract}
While lung cancer is the second most diagnosed form of cancer in men and women, a sufficiently early diagnosis can be pivotal in patient survival rates. Imaging-based, or radiomics-driven, detection methods have been developed to aid diagnosticians, but largely rely on hand-crafted features which may not fully encapsulate the differences between cancerous and healthy tissue. Recently, the concept of discovery radiomics was introduced, where custom abstract features are discovered from readily available imaging data. We propose a novel evolutionary deep radiomic sequencer discovery approach based on evolutionary deep intelligence. Motivated by patient privacy concerns and the idea of operational artificial intelligence, the evolutionary deep radiomic sequencer discovery approach organically evolves increasingly more efficient deep radiomic sequencers that produce significantly more compact yet similarly descriptive radiomic sequences over multiple generations. As a result, this framework improves operational efficiency and enables diagnosis to be run locally at the radiologist's computer while maintaining detection accuracy. We evaluated the evolved deep radiomic sequencer (EDRS) discovered via the proposed evolutionary deep radiomic sequencer discovery framework against state-of-the-art radiomics-driven and discovery radiomics methods using clinical lung CT data with pathologically-proven diagnostic data from the LIDC-IDRI dataset.  The evolved deep radiomic sequencer shows improved sensitivity (93.42\%), specificity (82.39\%), and diagnostic accuracy (88.78\%) relative to previous radiomics approaches.
\end{abstract}


{\noindent \footnotesize\textbf{*} Mohammad Javad Shafiee, \linkable{mjshafiee@uwaterloo.ca} }

\begin{spacing}{2}   

\section{Introduction}
\label{Introduction}
Lung cancer is the second most diagnosed form of cancer in men and women after prostate cancer and breast cancer, respectively. In 2016, lung cancer accounted for an estimated 158,080 deaths (approximately $27\%$ of cancer deaths) and 224,390 new cases in Americans~\cite{ACS2016}. Similarly, lung cancer accounted for an estimated 20,800 deaths (approximately $26\%$ of cancer deaths) and 28,400 new cases in Canadians~\cite{CCS2016}. Early detection of lung cancer can significantly impact the patient survival rate, making efficient and reliable lung cancer screening methods crucial.

Imaging-based cancer detection or radiomics-driven methods have recently grown in popularity to help streamline the cancer screening process and increase diagnostic consistency. Referring to the extraction and analysis of large amounts of quantitative features from medical imaging data, radiomics~\cite{Lambin2012} allows for the creation of a high-dimensional abstract feature space that can be utilized for cancer detection via the detailed characterization of cancer phenotypes. The prognostic potential of radiomics has previously been demonstrated in studies on lung and head-and-neck cancer patients~\cite{Aerts2014,Gevaert2012}. Aerts et al.~\cite{Aerts2014} introduced a comprehensive study spanning over 1000 patients across seven datasets to demonstrate the application of radiomics towards differentiating between tumour phenotypes, indicating clinical and prognostic implications. In addition, radiomics has shown promise in combination with multi-parametric magnetic resonance imaging for breast cancer detection~\cite{Maforo2015} and prostate cancer detection~\cite{Khalvati2015,Cameron2015}.

Radiomics-driven methods have previously been developed for malignant lung nodule detection using computed tomography (CT) images~\cite{Anirudh2016, Orozco2015, Shen2015, Shen2017}. Anirudh et al.~\cite{Anirudh2016} used weakly labelled lung data from the SPIE-LUNGx dataset to train a 3D convolutional neural network (CNN) and generate radiomic sequences for lung nodule detection. In contrast, Orozco et al.~\cite{Orozco2015} generated wavelet-based radiomic sequences and demonstrated the effectiveness of wavelet-based features using a subset of images from the early lung cancer action project (ELCAP) and lung image database consortium (LIDC) datasets.

Shen et al.~\cite{Shen2015} proposed multi-scale convolutional neural networks (MCNN), a hierarchical framework for extracting discriminative features from lung nodules. Specifically, the framework comprises of alternating, stacked layers, and uses multi-scale nodule patches to learn class-specific features. More recently, Shen et al. extended their previous work to malignancy suspiciousness classification~\cite{Shen2017}. In addition, the extension simplified the training process via a multi-crop pooling architecture. An important aspect of these aforementioned radiomics-driven methods is that they leverage radiologist-driven nodule annotations for predicting the malignancy of lung nodules, rather than using pathology-proven data.

There are relatively few radiomics-driven methods that perform lung cancer detection using pathology-proven diagnostic data~\cite{Kumar2015, Shen2016}. Kumar et al.~\cite{Kumar2015} introduced an unsupervised deep autoencoder for feature extraction with a binary decision tree classifier for lung nodule classification. Shen et al.~\cite{Shen2016} proposed a domain-adaptation framework for lung nodule malignancy prediction; more specifically, Shen et al. propose CNN-MIL for learning transferable patient-level malignancy knowledge, which combines a convolutional neural network (CNN) model with a multiple instance learning model (MIL).

\begin{figure*}[t]
	\centering
	\includegraphics[width = \linewidth]{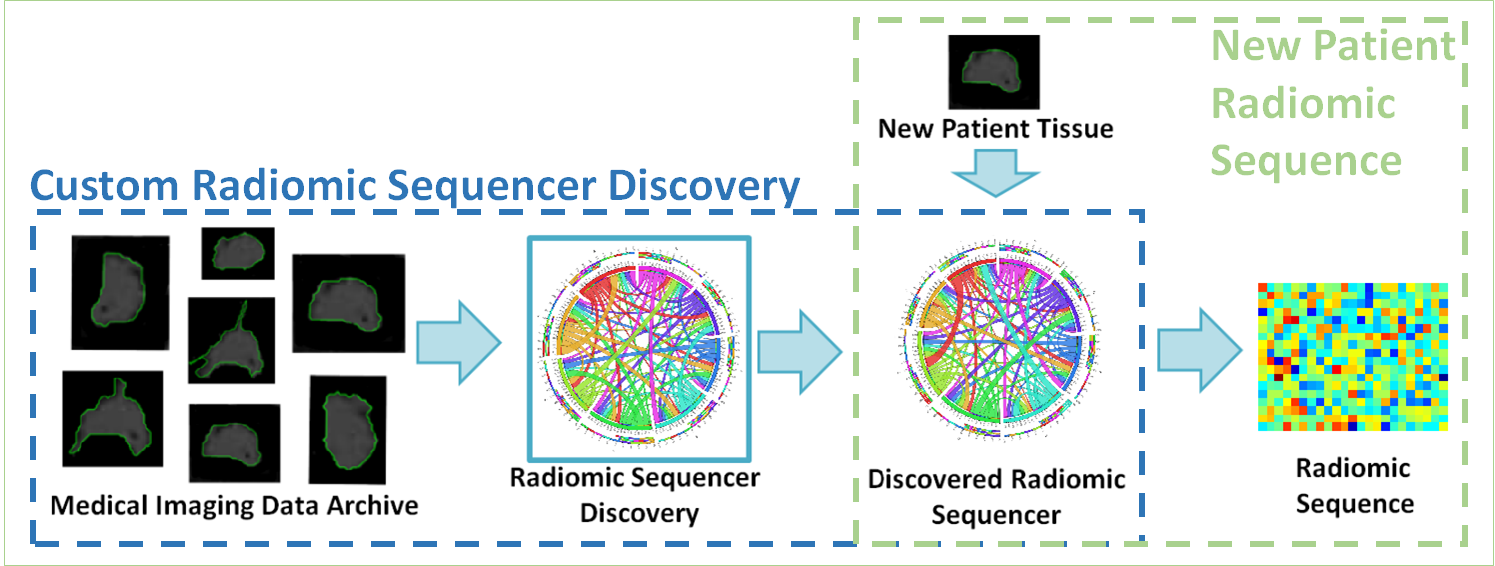}
	\caption{Overview of the discovery radiomics framework for cancer detection. A custom radiomic sequencer is discovered via past medical imaging data; for new patients, radiomic sequences of abstract imaging-based features are generated for quantification and analysis.}
	\label{fig_DiscRad}
\end{figure*}

Recently, the concept of \textit{discovery radiomics} was introduced where notions of using pre-defined, hand-crafted features for cancer detection are bypassed in favour of radiomic sequencers that produce abstract imaging-based features that are discovered directly from the wealth of readily-available medical imaging data. This allows for custom-tailored features to be discovered that can better characterize cancerous tissue and distinguish cancer phenotypes relative to conventional features. Discovery radiomics has shown promising results for both prostate cancer~\cite{Karimi2017} and lung cancer~\cite{Kumar2015,Shafiee2015, Kumar2017} detection.

A number of different radiomic sequencers have been proposed within the discovery radiomics framework for the purpose of lung cancer detection.  Kumar et al.~\cite{Kumar2015} introduced the notion of deep autoencoding radiomics sequencers (DARS), which comprises of a deep autoencoder architecture.  Shafiee et al.~\cite{Shafiee2015} proposed deep radiomics sequencers based on a deep convolutional StochasticNet~\cite{Shafiee2016} architecture, referred to as StochasticNet sequencers. More recently, Kumar et al.~\cite{Kumar2017} leveraged deep radiomic sequencers built upon a deep convolutional neural network architecture.

While diagnostically powerful, the discovered radiomic sequencers were both computationally expensive and memory intensive, which could make it difficult for on-site clinical deployment and would require the transfer of patient information to more powerful cloud computing leading to patient privacy concerns.  To mitigate computational requirements and increase operating efficiency, we propose a novel evolutionary deep radiomic sequencer discovery framework for discovering more efficient yet powerful deep radiomic sequencers.  Using the concept of \textit{evolutionary deep intelligence}~\cite{shafiee2016deep,shafiee2017evolution} to mimic biological evolution mechanisms, the proposed \textit{evolutionary deep sequencer discovery} process discovers progressively more efficient yet diagnostically powerful deep radiomic sequencers over multiple generations.  The resulting evolved deep radiomic sequencers (EDRS) that are not only significantly more efficient, thus making them more suitable for on-site clinical deployment, but can provide improved diagnostic performance compared to existing deep radiomic sequencers.

\section{Methods}
\label{Methods}
In this section, we will first discuss the concepts behind discovery radiomics and evolutionary deep intelligence.  We will then present the proposed evolutionary deep sequencer discovery approach in detail.

\subsection{Discovery Radiomics}
The idea behind discovery radiomics can be described as follows (see Figure~\ref{fig_DiscRad}). Given past radiology data and corresponding pathology-verified radiologist tissue annotations from a medical imaging data archive (i.e., provided by Cancer Imaging Archive~\cite{Armato1, Armato2} consists of diagnostic and lung cancer screening thoracic computed tomography), the radiomic sequencer discovery process learns a radiomic sequencer that can extract highly customized radiomic features (which we will refer to as a radiomic sequence) that are tailored for characterizing unique tissue phenotype that differentiate cancerous tissue from healthy tissue.  The discovered radiomic sequencer can be applied to a new patient data to extract the corresponding radiomic sequence for cancer screening and diagnosis purposes.

As discussed earlier, one of the key limitation of previously proposed deep radiomic sequencers~\cite{Kumar2015,Kumar2017} for the purpose of lung cancer detection is that, while diagnostically powerful, both computationally expensive and memory intensive. They usually utilize very deep neural network architectures with a large number of parameters (i.e., which needs a large amount of memory to store) such that a huge set of arithmetic operations are required to generate the radiomic sequence and as a result it needs a fair amount of time to produce the results. This could make it difficult for on-site clinical deployment and would require the transfer of patient information to more powerful cloud computing leading to patient privacy concerns.  To mitigate computational requirements and increase operating efficiency to enable on-site clinical deployment, we will leverage the concept of \textit{evolutionary deep intelligence}~\cite{shafiee2016deep,shafiee2017evolution} to discover highly efficient deep radiomic sequencers that still provide strong diagnostic performance.

\subsection{Evolutionary Deep Intelligence}
Prior to describing the proposed evolutionary deep sequencer discovery approach, it is first important to discuss the idea behind evolutionary deep intelligence.  First introduced by Shafiee et al.~\cite{shafiee2016deep}, the general idea is to synthesize progressively more efficient deep neural networks over multiple generations. The evolution of deep neural networks is modeled in a probabilistic manner, where the architectural traits of ancestor networks are encoded by a probabilistic DNA. The probabilistic DNA is utilized to mimic biological heredity, and new offspring networks are synthesized stochastically based on this probabilistic model. Each synaptic connectivity is modeled by a probability distribution based on the corresponding weight magnitude in the ancestor network such that the strength of the weight determines the probability of  each synapse to be  connected in the offspring network. To close the cycle of evolution, environmental factors are applied to the model to mimic random mutation and natural selection. The environmental factor is combined with the probabilistic DNA to enforce how the random mutation should be applied (i.e., what  the rate of mutation of synaptic connectivity should be in the offspring network architecture). Loosely speaking, when the environmental factor forces the offspring network architectures to be smaller than their ancestor, this causes to decrease the chance of each synaptic connectivity to be synthesized in the offspring network such that weaker synaptic connectivity in the ancestor network will have lower chance to be connected in the offspring network. At each generation, the offspring network (which is more efficient than its parent) is then trained to refine its modeling capabilities and maximize its modeling accuracy.

\begin{figure*}
	\centering
	\includegraphics[width = \linewidth]{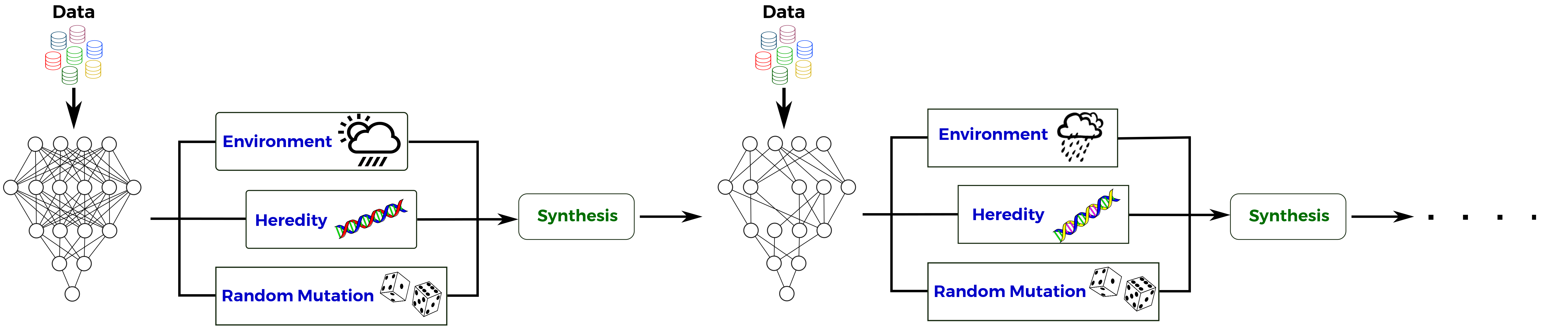}
	\caption{Evolutionary deep intelligence framework; The heredity is encoded by a probabilistic DNA modeling the architectural traits which should be carried to the next generation. The environmental conditions simulate the factors the must be considered to synthesized an offspring network. The evolutionary approach is repeated over multiple generations until all conditional are satisfied by latest generation.   }
	\label{fig_EvoNet}
\end{figure*}

Figure~\ref{fig_EvoNet} demonstrates the evolution process visually. As seen, the evolution is initialized using a known network structure as the first generation. The network is trained based on the available training data and the weights associated with each synaptic strength are computed. The underlying heredity of the network (i.e., as the parent network) is encoded by the probabilistic DNA which is modeled based on the synaptic strengths. The environmental factors are then formulated into the model to account for the requirements needed to be satisfied by the offspring network. The offspring network is then synthesized by taking advantage of random mutation to diversify the offspring network from its ancestors. This process is repeated until all requirements are satisfied by the latest offspring network. Given its ability to produce progressively more efficient yet powerful deep neural networks, we are motivated to leverage the ideas behind evolutionary deep intelligence within the discovery radiomics framework to discover highly efficient yet diagnostically accurate deep radiomic sequencers for the purpose of lung cancer detection.

\subsubsection{Evolutionary Deep Radiomic Sequencer Discovery}

\begin{figure}[!t]
	\centering
	\includegraphics[width = \linewidth]{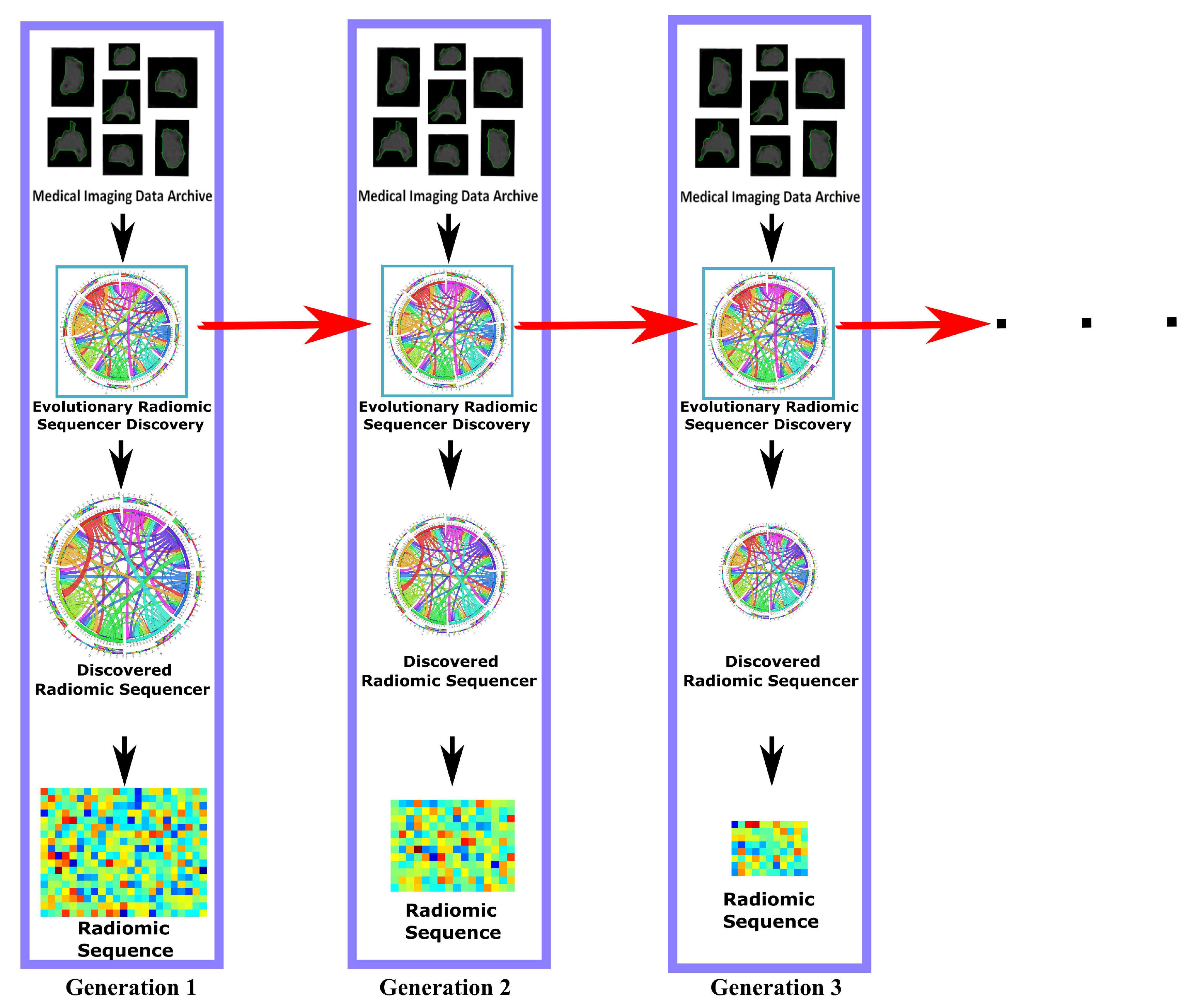}
	\caption{Evolutionary deep radiomic sequencer discovery to synthesize optimized radiomic sequencer from an archive of medical images (in this study, lung nodule CT images). At each generation, the past radiomic sequencer and archive of medical images are used by the evolutionary deep radiomic sequencer discovery process to synthesize a more efficient radiomic sequencer. As shown, the size of parameters of  radiomic sequencer is decreased over generation resulting to a more concise radiomic sequence to describe the input radiology image.}
	\label{fig_DiscoveryEvoNet}
\end{figure}

Motivated to leverage evolutionary deep intelligence within the discovery radiomics framework, we introduce an evolutionary deep radiomic sequencer discovery process for discovering deep radiomic sequencers. As seen in Figure~\ref{fig_DiscoveryEvoNet}, the evolutionary deep radiomic sequencer discovery  framework discovers a more optimal deep radiomic sequencer generation by generation and, as a result, the generated radiomic sequence at each generation is more concise compared to radiomic sequences generated by previous deep radiomic sequencers in past generations.

\begin{figure*}
	\centering
	\includegraphics[width = 0.6\linewidth]{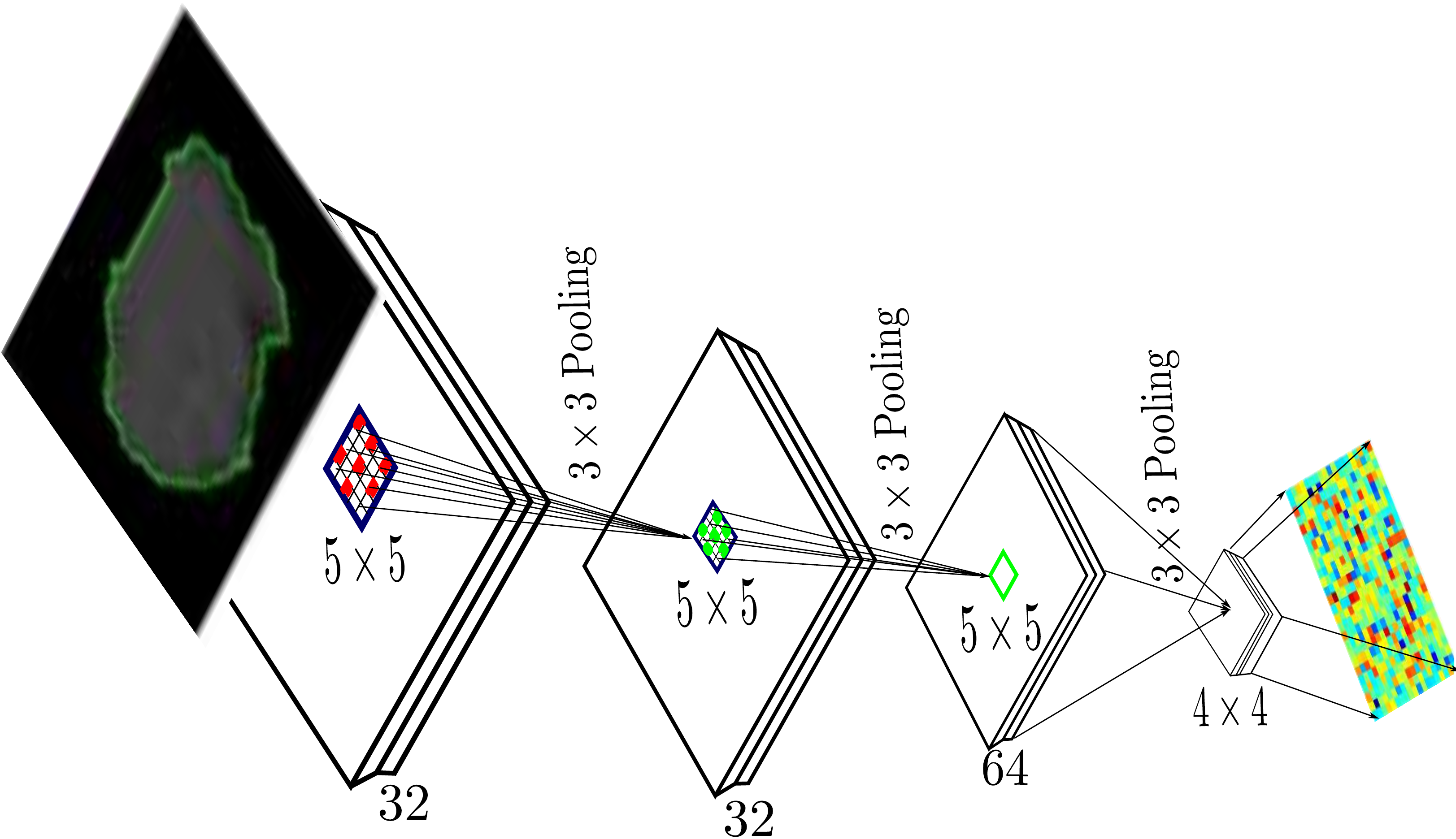}
	\caption{Deep radiomic sequencer based on deep convolutional neural network architecture. The input to the deep convolutional neural network  is a suspicious region of a CT image. The architecture of the original ancestor network (i.e., generation 1 in Figure~\ref{fig_DiscoveryEvoNet}) is a Lenet5 network architecture where it has 32 @ $5\times5$ filters in the first and second layers, 64 @ $5\times5$ filters in the third layers and 64 @ $4\times4$ in the last layer. The output of last layer generate the radiomic sequence with 1024 feature length.       }
	\label{fig_filter}
\end{figure*}

The methodology behind the proposed framework can be described as follows.  Inspired by~\cite{shafiee2016evolutionary, shafiee2017evolution}, let the deep radiomic sequencer be modeled as $\mathcal{H}(N,\mathbb{S})$, denoting a network architecture with the set of neurons $N$ and set of synaptic connectivities $\mathbb{S}$.  In this study, we will utilize a deep convolutional  neural network (CNN) architecture for the deep radiomic sequencer (see Figure~\ref{fig_filter}).  The structural information of a deep radiomic sequencer at generation $g$ can be encoded by $\mathbb{S}_g$. $W_{g-1}$ is the set of weights that encode the strength associated with each synapse in the network at generation $g-1$ where a synaptic weight of zero indicates that the associated synapse is not connected. It should be noted that $W_{g-1}$ can therefore encode the structural information, $\mathbb{S}_{g-1}$, of network at generation $g-1$. As a result, it is possible to reformulate  $P(\mathcal{H}_g|\mathcal{H}_{g-1})$ as $P(\mathbb{S}_g|W_{g-1})$  without any loss on modeling accuracy. Thus, the probabilistic DNA of a deep radiomic sequencer at generation $g$ is formulated as $P(\mathbb{S}_g|W_{g-1})$, such that at each generation $g$ the structure of the sequencer $\mathbb{S}_g$ is synthesized given the trained weights of the sequencer of the previous generation $W_{g-1}$.

The genetic encoding scheme (i.e., probabilistic DNA) can be formulated in different ways to favor special requirements needed to be applied when the new offspring deep radiomic sequencers are synthesized. For promoting computational efficiency and compactness, $P(\mathbb{S}_g|W_{g-1})$ is modeled such that it promotes the formation of a particular cluster of synapses while considering the synthesis of each individual synapse in the offspring deep radiomic sequencer as well~\cite{shafiee2017evolution}:
\begin{align}
	P(\mathbb{S}_g|W_{g-1}) = \prod_{c \in C} \Big [ P(S^c_g|W_{g-1}) \cdot \prod_{i \in c} P(s_g^i|w_{g-1}^i)\Big]
	\label{eq:cluster_driven}
\end{align}
where $P(S^c_g|W_{g-1})$ promotes the synthesis of a particular cluster of synapses $c$, $S^c_g \subset \mathbb{S}_g$ , given the weights of the network at generation $g-1$ and $P(s_g^i|w_{g-1}^i) $ is the probability that synapse $s_g^i \in S^c$ will be synthesized in the offspring deep radiomic sequencer at generation $g$.

A cluster of synapses can be defined and represented based on different factors, such as faster run-time of the offspring radiomic sequencer on a parallel computing device or decreased storage requirements relative to its ancestor sequencer. However, the main advantage of~\eqref{eq:cluster_driven} is that the $P(S^c_g|W_{g-1})$ not only favors strong synapses which are more effective in maintaining a high modeling accuracy, it promotes the persistence of clusters of synapses in the offspring deep radiomic sequencer which can extract more discriminative features, resulting in a sequencer that can model the problem more accurately.  Here we define a set of synapses constructing a filter in each convolutional layer as a cluster of synapses in the network structure of deep radiomic sequencer. As shown in Figure~\ref{fig_filter}, each filter in a convolutional layer is responsible for producing one output channel of the layer.  By extending this definition to all convolutional layers in the radiomic sequencer, the length of the radiomic sequence varies over the generations as the number of filters in the last layer determines the actual length of the radiomic sequence.

The probabilistic DNA, $P(S_g|W_{g-1})$, is combined with the environmental factor model $\mathcal{F}(\mathcal{E})$ to mimic natural selection, such that the offspring deep radiomic sequencer for the next generation is comprised of stochastically selected synapses or clusters of synapses. The environmental factor simulates the conditions which the offspring networks should be adapted for. For example, if in the new environment the offspring network should be faster in computation, the environmental factors enforce the offspring network to be synthesized with less number of filters for decreasing the processing time. The environmental factors also can reflect what is the situation in terms of memory availability and the host hardware in which the computation will be done; as a result the synthesized offspring network architectures adapts itself to these environmental factors to be able to survive.    
 The probabilistic model of the network structure, $P(\mathcal{H}_g)$, at generation $g$ can be formulated as
\begin{align}
	P(\mathcal{H}_g) = \mathcal{F}(\mathcal{E}) \cdot P(\mathbb{S}_g|W_{g-1})
\end{align}
where $\mathcal{F}(\mathcal{E})$ quantitatively encodes the environmental conditions, and  the offspring deep radiomic sequencer structures must adapt to them to survive over generations. As mentioned before, the goal here is to synthesize a deep radiomic sequencer with fewer parameters while preserving the modeling accuracy; therefore, the environmental factor model, $\mathcal{F}(\mathcal{E})$, favors the formation of a deep radiomic sequencer with fewer parameters and increased efficiency over the generations. This property is applied via a cluster-based encoding scheme which decreases the number of filters of different layers over generations:
\begin{align}
	P(\mathcal{H}_g) = \prod_{c \in C} \Big [\mathcal{F}_c(\mathcal{E}) \cdot P(S^c_g|W_{g-1}) \Big].
\end{align}
More specifically, the environmental factor $\mathcal{F}_c(\mathcal{E})$ is formulated such that the offspring radiomic sequencer is limited to 80\% of the total number of synapses in its direct ancestor sequencer.

\section{Results}
\label{Results}
\subsection{Experimental Setup}
The proposed evolutionary deep radiomic sequencer discovery framework was examined using the pathology-proven subset of the LIDC-IDRI~\cite{Armato1, Armato2} dataset, and was compared to state-of-the-art methods. In this section, the configuration of dataset, the underlying network architecture of the discovered radiomic sequencers, and the competing methods are explained.

\subsubsection{Lung Dataset}
In this study, we used the subset of the LIDC-IDRI~\cite{Armato1, Armato2} dataset that had corresponding pathology-proven diagnostic data. The dataset is a public dataset  provide by Cancer Imaging Archive~\cite{Armato1, Armato2} consists of diagnostic and lung cancer screening thoracic computed tomography (CT) scans with marked-up annotated lesions. The CT images were captured using a broad range of scanner models from different manufacturers by applying the following tube peak potential energies for acquiring the scans: 120$kV$ ($n = 818$), 130$kV$ ($n=31$), 135$kV$ ($n=69$), and 140$kV$ ($n=100$). A subset of 93 patient cases which have definite diagnostic results was selected from the LIDC-IDRI. While pathology data was used to generate labels for the CT images in the LIDC-IDRI dataset, the pathology data itself was not available for comparison and labels were provided on a nodule basis. Using data augmentation, an enriched dataset of 42,340 lung lesions was obtained via the rotation of each malignant and benign lesion by 45$^\circ$ and 10$^\circ$ increments, respectively. The proposed method is examined by a 10-fold cross-validation approach where 9 out of 10 folds  of patient cases (subset of patient cases) are used in the training while the other fold (subset of patient cases)  is utilized as test samples and the results are reported based on the average performance of 10 trials.

\subsubsection{Network Architecture (Lenet5)}
The deep neural network architecture of the original, first generation  radiomic sequencer used in this study builds upon the Lenet5 architecture~\cite{lecun1998}. The radiomic sequencer is comprised of three convolutional layers: $c_1: 3\times3$, $c_2: 5\times5$, and $c_3: 3 \times 3$, where the first layer consists 32 filters, the second layer has 32 filters, and the last layer has 64 filters. The radiomic sequence generated by  the original, first generation radiomic sequencer has a length of $16 \times 64$, and is the input into two fully-connected layers ($f_1: 64$ and $f_2: 2$) to classify each input as cancerous or benign.

\subsubsection{Competing Frameworks}
The proposed evolutionary deep radiomic sequencer discovery was evaluated using the enriched dataset and quantitatively compared to four state-of-art radiomics-driven approaches~\cite{Kumar2015, Shen2016, Shafiee2015, Kumar2017}.

Kumar et al.'s deep autoencoding radiomic sequencer (DARS)~\cite{Kumar2015} uses a five layer de-noising autoencoder trained by L-BFGS with 30 iterations and a batch size of 400, as suggested by past work~\cite{Ngiam2011}; a 200 dimension feature vector is extracted from the fourth layer and paired with a binary decision tree classifier. Shen et al.'s proposed convolutional neural network multiple instance learning (CNN-MIL)~\cite{Shen2016} is composed of three concatenated convolutional layers, each with 64 convolutional kernels of size $3 \times 3$. Each convolutional layer is followed by a rectified linear unit and a max-pooling layer ($4 \times 4$ pooling window in the first layer, and $2 \times 2$ in the subsequent layers), and two fully-connected layers are used to determine nodule malignancy.  Shafiee et al.'s StochasticNet radiomic sequencer (SNRS)~\cite{Shafiee2015} is constructed using three stochastically-formed convolutional layers of 32, 32, and 64 receptive fields, respectively. Each receptive field is $5 \times 5$ in size, and is part of a random graph realization with a uniform neural connectivity probability of 0.5. Similarly, Kumar et al.'s discovered radiomic sequencer (DRS)~\cite{Kumar2017} is comprised of three convolutional sequencing layers of 20, 50, and 500 receptive fields, respectively, each of size $3 \times 3$.

\subsection{Experimental Results}
The proposed evolutionary deep radiomic sequencer discovery process was performed through 11 generations where in each generation, the environmental factor restricts the offspring radiomic sequencer to 80\% of the total number of synapses in its direct previous network. By using this environmental factor, the number of parameters in the deep neural network of radiomic sequencer is decreased generation by generation, allowing for the generated radiomic sequences to be more compact over generations. Decreasing the number of parameters in the sequencer is important as it affects the generalizability of the sequencer such that a more generalized sequencer is less likely to be over-trained to the training data and can perform more accurately in the evaluation step.

The performance of the proposed framework is examined in a 10-fold cross validation approach where 9 out of 10 subsets of the data are used in the training step while the 10th subset is used to evaluate the model. This training and testing process is repeated over all permutations of the training and testing subsets. The cross validation approach is combined with evolutionary deep intelligence, where in each validation step, the radiomic sequencers are synthesized generation by generation with the same training dataset and validated with the same testing data.

Table~\ref{tab_genInfo} shows the average performance of the proposed framework over 11 generations. As seen, by moving generation by generation, the number of filters used in the radiomic sequencer is decreased and the length of the radiomic sequence is correspondingly shortened. However, the performance of the radiomic sequencers improves over generations which demonstrates the increase in the generalizability of the models through generations. As seen, the performance of evolved radiomic sequencers (i.e., sensitivity, specificity and accuracy) increases after first generation and it reaches a stable point (e.g., generation 7). However evolving the radiomic sequencer after 7th generation can improve the compactness of the radiomic sequences.

Table~\ref{tab_genInfo} demonstrates that the specificity of the radiomic sequencers increases when the sequencers are evolved over generations, which is a good indication of generalizability of the final model. It is worth noting that in lung cancer classification, improving the specificity is challenging~\cite{toyoda2008sensitivity} and increasing the specificity while maintaining a reasonable sensitivity is highly desirable.

Table~\ref{tab_genInfo} also shows that the evolved radiomic sequencers can perform better in terms of sensitivity compared to the first generation original ancestor radiomic sequencer, resulting in a model with higher accuracy. As mentioned before, one of the important obstacles in using a deep neural network as the underlying architecture for a radiomic sequencer is the efficiency of the underlying deep neural network. As seen, the average number of filters constructing the radiomic sequencer is decreased over generations, indicating that the efficiency of the radiomic sequencer is increasing generation by generation. It is also worth noting that the number of filters of a deep neural network determines the number of parameters need to be computed in one forward pass of the network to compute the final prediction; therefore, decreasing this number can increase the efficiency of the radiomic sequencer.

\begin{figure*}
	\centering
	\includegraphics[width = 0.9\linewidth]{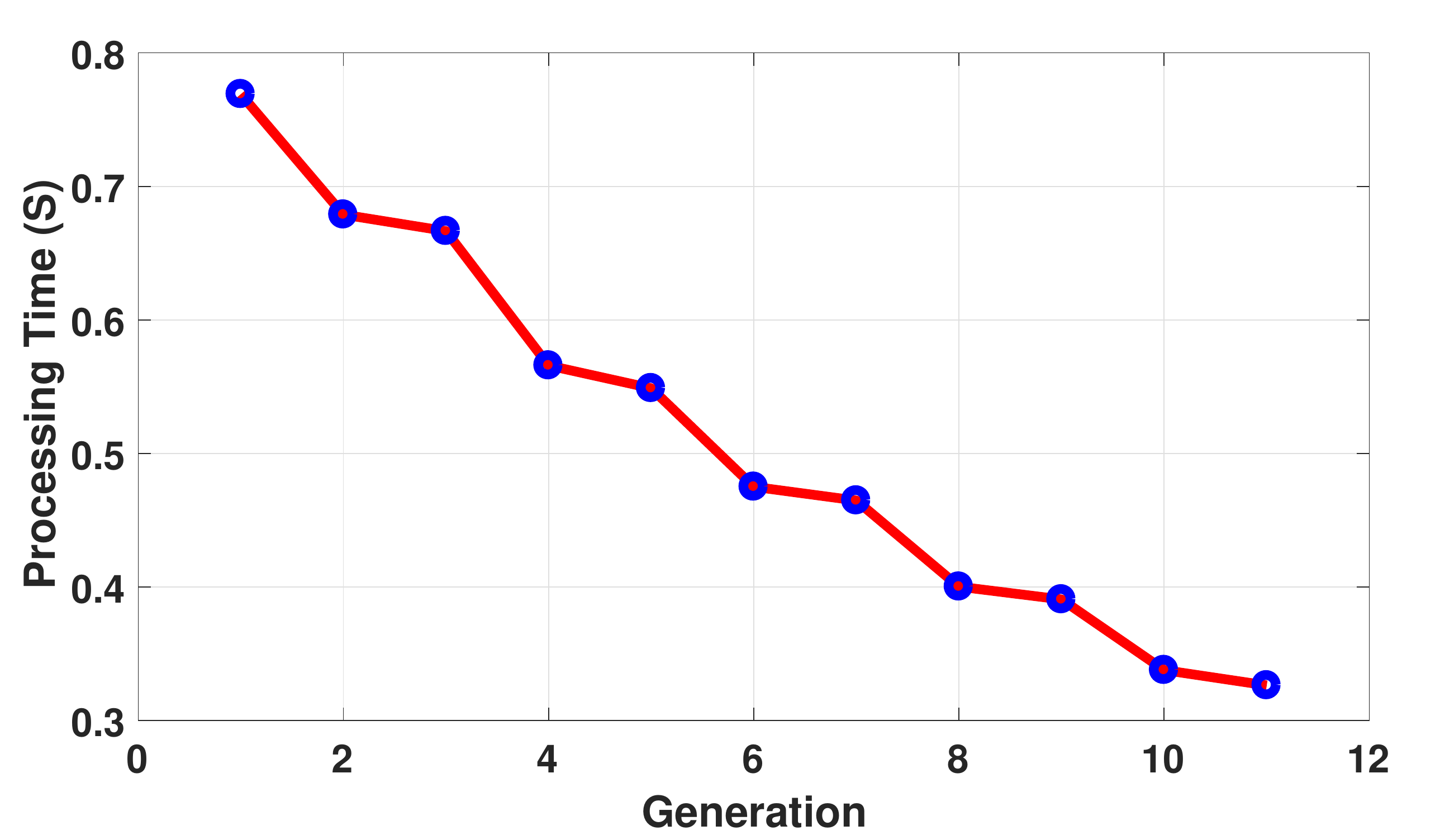}
	\caption{Running time evaluation of synthesized radiomic sequencer at each generation is evaluated by 1500 sample inputs. As seen the subsequent radiomic sequencers perform faster than their ancestors which shows the efficiency of the synthesized sequencer discovered via the proposed evolutionary deep radiomic sequencer discovery process.   }
	\label{runtime}
\end{figure*}

To evaluate the efficiency of the synthesized radiomic sequencers, the running time computation of the sequencers are examined at each generation with 1500 sample inputs. Figure~\ref{runtime} demonstrates the running time performance  of synthesized radiomic  sequencers through generations. As seen the subsequent generations performs faster than their ancestors which shows the efficiency of the proposed evolutionary deep intelligence framework.  

\begin{table}		
	\begin{center}
		\caption{Radiomic sequence lengths and the modeling accuracies over generations. ``A.N.F" stands for average number of filters  in the sequencer and ``R.S.L'' column  represents the average length of the radiomic sequence at each generation. Since the numbers are averaged over 10 folds of evaluation they are reported with one floating point precision. As seen while the radiomic sequences become more compact over generations, the modeling accuracy, sensitivity, and specificity are increasing.}
		\setlength{\tabcolsep}{8pt}
		\label{tab_genInfo}
		\begin{tabular}{|l|ccccc|}
			\hline
			~&\bf A.N.F. & \bf R.S.L. &\bf Sensitivity &\bf Specificity &\bf Accuracy \\ \hline
			\bf Gen.1& 194.0	&3104.0	&0.8786&	0.7570&	0.8255\\
			\bf Gen.2&180.4&	2886.4&	0.9156&	0.7788&	0.8590\\
			\bf Gen.3&171.0&	2736.0&	0.9305&	0.8063&	0.8795\\
			\bf Gen.4&161.1&	2577.6&	0.9276&	0.8062&	0.8812\\
			\bf Gen.5&150.9&	2414.4&	0.9311&	0.8109&	0.8845\\
			\bf Gen.6&142.6&	2281.6&	0.9295&	0.8125&	0.8834\\
			\bf Gen.7&135.1&	2161.6&	\bf 0.9390&	0.8105&	0.8898\\
			\bf Gen.8&125.8&	2012.8&	0.9341&	0.8129&	0.8879\\
			\bf Gen.9&118.5&	1896.0&	0.9384&	0.8169&	0.8917\\
			\bf Gen.10&111.2&	1779.2&	0.9385&	0.8107&	\bf 0.8901\\
			\bf Gen.11&104.5&	1672.0&	0.9342&	\bf 0.8239&	0.8878\\	\hline
		\end{tabular}		
	\end{center}		
\end{table}

Decreasing the number of filters in the model decreases the length of the radiomic sequence. As shown in Table~\ref{tab_genInfo}, the length of the radiomic sequence is shortened generation by generation and the length of radiomic sequence in the last generation is about half size of the radiomic sequence of the first generation, demonstrating that it is possible to increase the concision of the radiomic sequence while simultaneously increasing the modeling accuracy.

\begin{table}
	\begin{center}		
		\caption{Comparison with state-other-the-art methods for lung cancer classification. As seen the proposed EDRS framework outperforms other methods in sensitivity, specificity, and accuracy.}
		\setlength{\tabcolsep}{8pt}
		\label{tab:compInfo}
		\begin{tabular}{l|ccccc}
			\hline
			~&\bf Sensitivity &\bf Specificity &\bf Accuracy \\ \hline
			\bf DARS~\cite{Kumar2015}	& 0.8314 	& 0.2018	& 0.7501 \\
			\bf CNN-MIL~\cite{Shen2016}	& --		& --		& 0.7069 \\
			\bf SNRS~\cite{Shafiee2015}	& 0.9107 	& 0.7598	& 0.8449 \\
			\bf DRS~\cite{Kumar2017}	& 0.7906	& 0.7611	& 0.7752 \\
			\bf EDRS 	& \textbf{0.9342}	& \textbf{0.8239}	& \textbf{0.8878} \\ \hline
			\bf Last-Generation & 0.8893	& 0.7823	& 0.8355\\
			\hline
		\end{tabular}
	\end{center}
\end{table}

\begin{figure*}
	\centering
	\includegraphics[width = 0.9\linewidth]{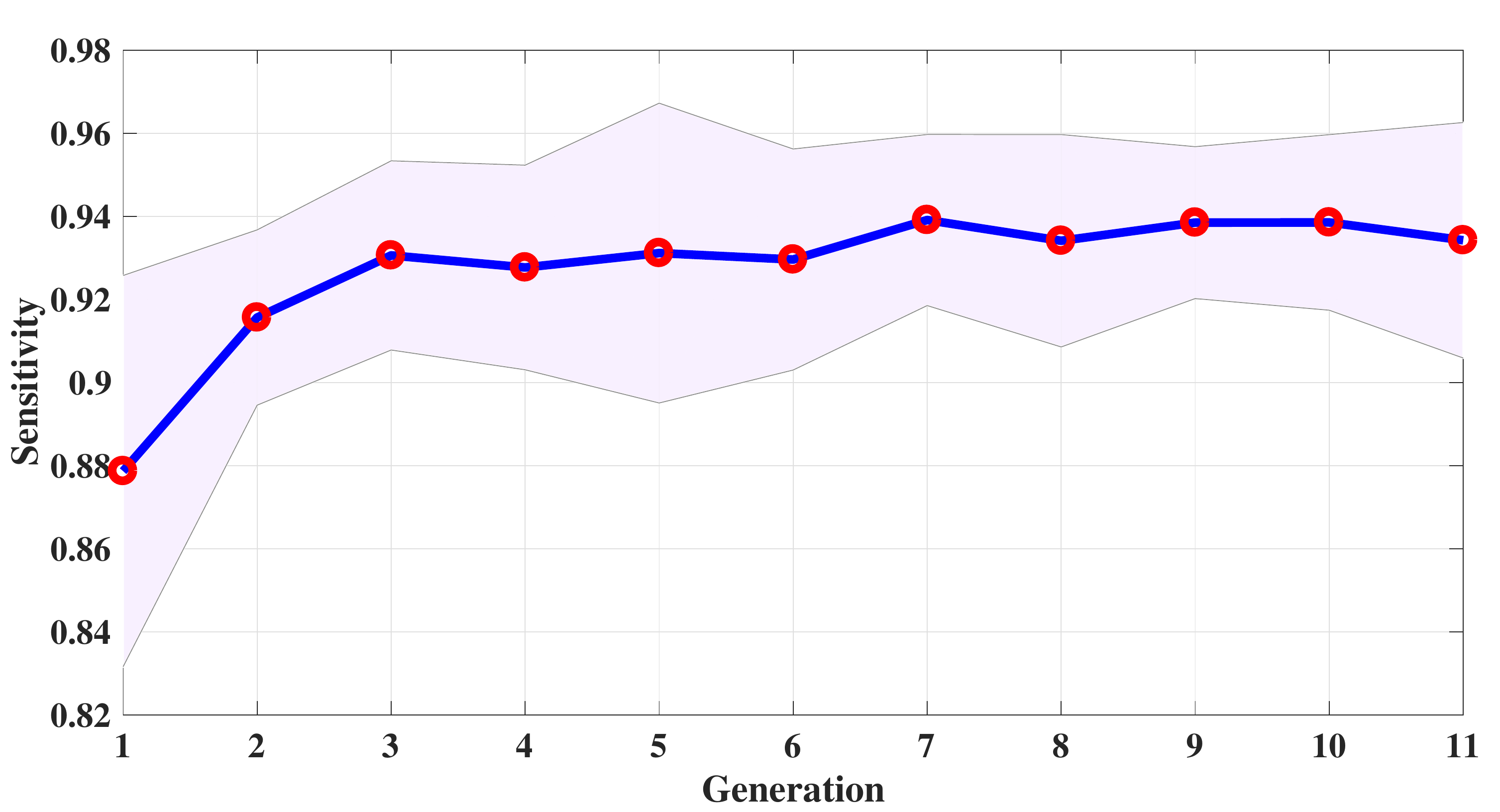}
	\caption{Sensitivity of the evolved radiomic sequencers over generations. The standard deviation of the models based on 10-fold cross validation is overlaid with purple margin. }
	\label{fig_Sen}
\end{figure*}

\begin{figure*}
	\centering
	\includegraphics[width = 0.9\linewidth]{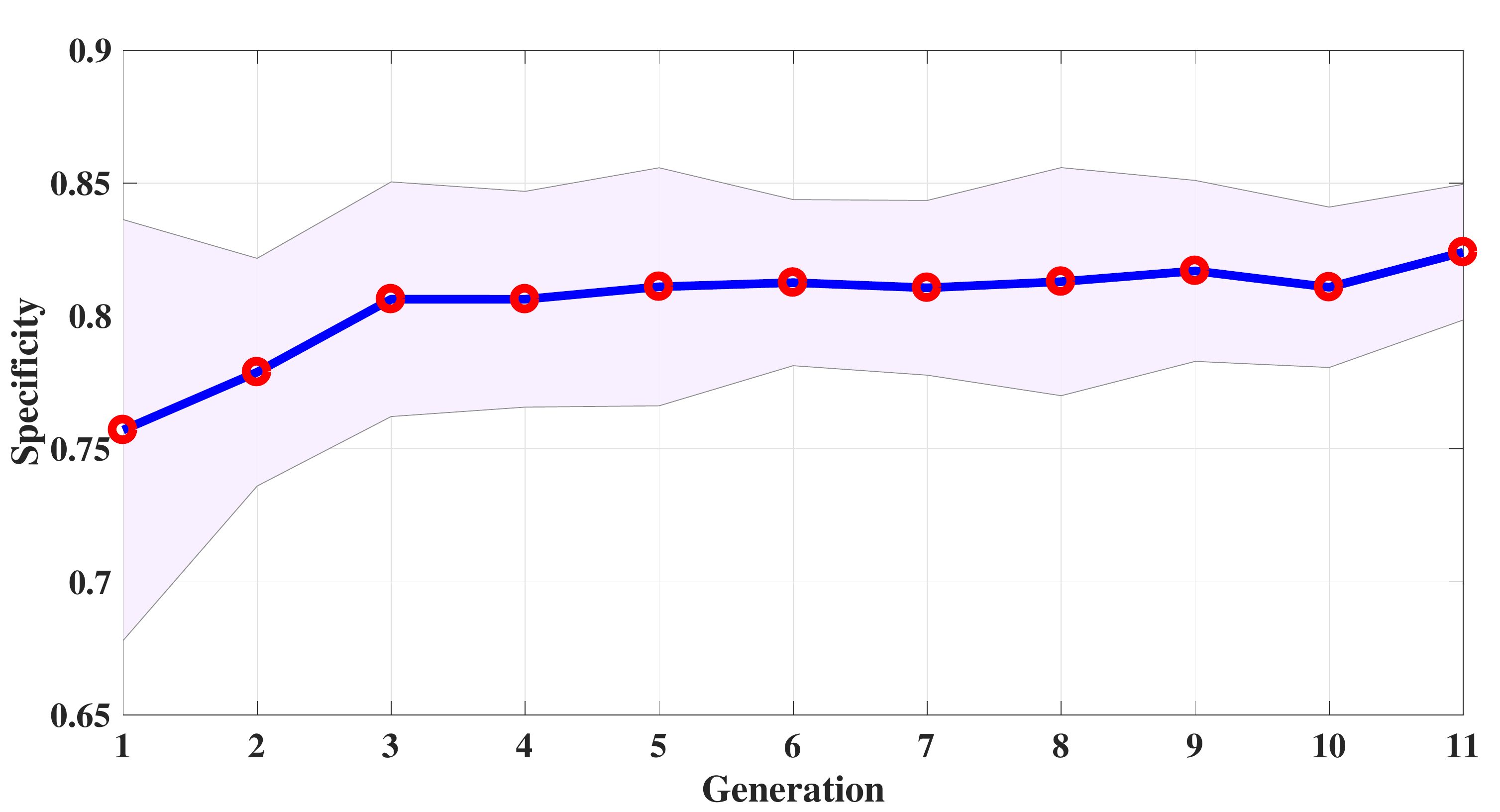}
	\caption{Specificity of the radiomic sequencer overlaid by their modeling standard deviation over generations. As seen, the generalizability of radiomic sequencer increase generation by generation as the standard deviation of modeling decreases. }
	\label{fig_Spc}
\end{figure*}

\begin{figure*}
	\centering
	\includegraphics[width = 0.9\linewidth]{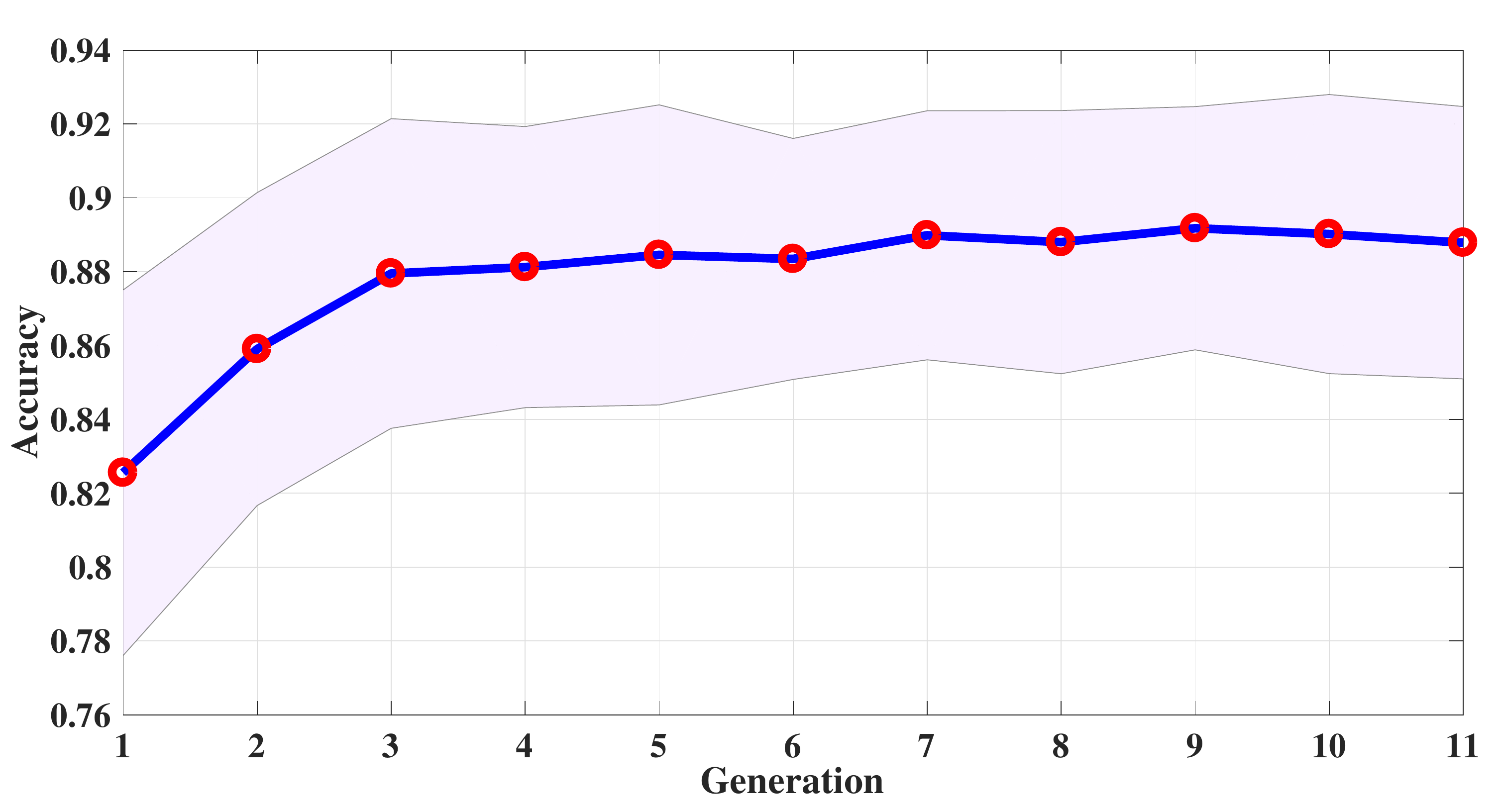}
	\caption{Radiomic sequencers modeling accuracy over generations. }
	\label{fig_Acc}
\end{figure*}

Figure~\ref{fig_Sen} demonstrates the sensitivity of the evolved radiomic sequencers overlaid with the standard deviation across different folds of cross validation over multiple generations. By evolving the radiomic sequencers generation by generation, the sensitivity increases while the standard deviation decreases (notice that the purple margin narrows over generations). This is another indication of generalizability of the evolved radiomic sequencers as the variance of the models in different cross validation folds of evaluation decreases over generations.
This effect is more obvious in Figure~\ref{fig_Spc} as the standard deviation of the specificity measure is decreased generation by generation and as mentioned before, a more reliable specificity is highly desirable in lung cancer classification. Figure~\ref{fig_Acc} shows the same behavior of the modeling accuracy over generations.

As the last experimental result, Table~\ref{tab:compInfo} shows the comparison of the proposed framework (EDRS) with other state-of-the-art approaches. It should be noted that the statistics and modeling performances of other state-of-the-art frameworks are reported directly from Kumar et al.~\cite{Kumar2017} and Shafiee et al.~\cite{Shafiee2016}. As seen, the proposed radiomic sequencer in the discovery radiomics framework outperforms other state-of-the-art methods in sensitivity (93.42\%), specificity (82.39\%) and accuracy (88.78\%).  To demonstrate the effect of evolutionary deep intelligence framework on discovery radiomic sequencer, the final network architecture synthesized by the evolutionary deep intelligence framework is trained from scratch. The performance of this network (so-called Last-Generation) is compared with the result of evolutionarily deep intelligence approach. As seen in Table~\ref{tab_genInfo}, although the optimized network architecture synthesized by the evolutionary framework is utilized to train Last-Generation approach, the sequencer could not compete with the EDRS performance and could not gain the same accuracy level. 

\section{Conclusion}
\label{Conclusion}
In this paper, we proposed a new evolutionary deep radiomic sequencer discovery framework to better uncover more efficient yet powerful radiomic sequencers for the purpose of lung cancer classification. An evolutionary deep intelligence approach is incorporated within the discovery radiomics framework to evolve the underlying deep neural network architecture of the deep radiomic sequencer over multiple generations and discover a more efficient and generalized deep radiomic sequencer. The ultimate goal here is to synthesize a new deep neural network as the underlying core of radiomic sequencer with a fewer number of parameters which produce more concise radiomic sequences that can better capture the differences between healthy and cancerous lung tissue. Results show that by evolving  and discovering more efficient radiomic sequencers, the diagnostic accuracy can be increased. Experimental results demonstrate that the evolved deep radiomic sequencer discovered using the proposed evolutionary deep radiomic sequencer discovery approach can outperform other state-of-the-art radiomics-driven methods, achieving a sensitivity of 93.42\%, a specificity of 82.39\%, and an accuracy of 88.78\%. It has been showed in the literature that there is a direct relation between the number of parameters  and the need for training data. As a future work it is suggested to study the effect of evolutionary deep intelligence framework when a limited training data is available.

\vspace{2ex}
\noindent\textit{Disclosures}

\noindent The authors have no relevant financial interests in the manuscript and no other potential conflicts of interest to disclose.

\acknowledgments
This research has been supported by the Ontario Institute of Cancer Research (OICR), Canada Research Chairs programs, Natural Sciences and Engineering Research Council of Canada (NSERC), and the Ministry of Research and Innovation of Ontario.  The authors also thank Nvidia for the GPU hardware used in this study through the Nvidia Hardware Grant Program.


\bibliography{evodr}   

\begin{thebibliography}{10}

\bibitem{ACS2016}
{American Cancer Society}, ``{Cancer Facts \& Figures 2016},''  (2016).

\bibitem{CCS2016}
{Canadian Cancer Society}, ``{Prostate Cancer Statistics},''  (2016).

\bibitem{Lambin2012}
P.~Lambin, E.~Rios-Velazquez, R.~Leijenaar, {\em et~al.}, ``Radiomics:
  extracting more information from medical images using advanced feature
  analysis,'' {\em European Journal of Cancer} {\bf 48}(4), 441--446  (2012).

\bibitem{Aerts2014}
H.~J. W. L. e.~a. Aerts, ``{Decoding tumour phenotype by noninvasive imaging
  using a quantitative radiomics approach.},'' {\em Nat Commun} {\bf 45}(4)
  (2014).

\bibitem{Gevaert2012}
O.~Gevaert, J.~Xu, C.~Hoang, {\em et~al.}, ``{Non-small cell lung cancer:
  identifying prognostic imaging biomarkers by leveraging public gene
  expression microarray data},'' {\em Radiology} {\bf 2}(4), 387--96  (2012).

\bibitem{Maforo2015}
N.~Maforo, H.~Li, W.~Weiss, {\em et~al.}, ``Radiomics of multi-parametric
  breast mri in breast cancer diagnosis: A quantitative investigation of
  diffusion weighted imaging, dynamic contrast-enhanced, and t2-weighted
  magnetic resonance imaging,'' {\em Medical physics} {\bf 42}(6), 3213--3213
  (2015).

\bibitem{Khalvati2015}
F.~Khalvati, A.~Wong, and M.~A. Haider, ``Automated prostate cancer detection
  via comprehensive multi-parametric magnetic resonance imaging texture feature
  models,'' {\em BMC medical imaging} {\bf 15}(1), 27  (2015).

\bibitem{Cameron2015}
A.~Cameron, F.~Khalvati, A.~Wong, {\em et~al.}, ``Maps: A quantitative
  radiomics approach for prostate cancer detection,'' {\em IEEE Transactions on
  Biomedical Engineering} {\bf 63}(6), 1145--1156  (2015).

\bibitem{Anirudh2016}
R.~Anirudh, J.~J. Thiagarajan, T.~Bremer, {\em et~al.}, ``Lung nodule detection
  using 3d convolutional neural networks trained on weakly labeled data,'' in
  {\em SPIE Medical Imaging},  978532--978532, International Society for Optics
  and Photonics  (2016).

\bibitem{Orozco2015}
H.~M. Orozco, O.~Villegas, V.~G.~C. S{\'a}nchez, {\em et~al.}, ``Automated
  system for lung nodules classification based on wavelet feature descriptor
  and support vector machine,'' {\em Biomedical engineering online} {\bf
  14}(1), 9  (2015).

\bibitem{Shen2015}
W.~Shen, M.~Zhou, F.~Yang, {\em et~al.}, ``Multi-scale convolutional neural
  networks for lung nodule classification,'' 588--599  (2015).

\bibitem{Shen2017}
W.~Shen, M.~Zhou, F.~Yang, {\em et~al.}, ``Multi-crop convolutional neural
  networks for lung nodule malignancy suspiciousness classification,'' {\em
  Pattern Recognition} {\bf 61}, 663--673  (2017).

\bibitem{Kumar2015}
D.~Kumar, A.~Wong, and D.~A. Clausi, ``Lung nodule classification using deep
  features in ct images,'' {\em Computer and Robot Vision (CRV), 2015 12th
  Conference on} , 133--138  (2015).

\bibitem{Shen2016}
W.~Shen, M.~Zhou, F.~Yang, {\em et~al.}, ``Learning from experts: Developing
  transferable deep features for patient-level lung cancer prediction,'' in
  {\em International Conference on Medical Image Computing and
  Computer-Assisted Intervention},  124--131, Springer  (2016).

\bibitem{Karimi2017}
A.-H. Karimi, A.~G. Chung, M.~J. Shafiee, {\em et~al.}, ``Discovery radiomics
  via a mixture of deep convnet sequencers for multi-parametric mri prostate
  cancer classification,'' in {\em International Conference Image Analysis and
  Recognition},  Springer  (2017).

\bibitem{Shafiee2015}
M.~J. Shafiee, A.~G. Chung, D.~Kumar, {\em et~al.}, ``Discovery radiomics via
  stochasticnet sequencers for cancer detection,'' {\em NIPS Workshop on
  Machine Learning in Healthcare}   (2015).

\bibitem{Kumar2017}
D.~Kumar, A.~G. Chung, M.~J. Shafiee, {\em et~al.}, ``Discovery radiomics for
  pathologically-proven computed tomography lung cancer prediction,'' in {\em
  International Conference Image Analysis and Recognition},  Springer  (2017).

\bibitem{Shafiee2016}
M.~J. Shafiee, P.~Siva, and A.~Wong, ``Stochasticnet: Forming deep neural
  networks via stochastic connectivity,'' {\em IEEE Access} {\bf 4}, 1915--1924
   (2016).

\bibitem{shafiee2016deep}
M.~J. Shafiee, A.~Mishra, and A.~Wong, ``Deep learning with darwin:
  Evolutionary synthesis of deep neural networks,'' {\em arXiv preprint
  arXiv:1606.04393}   (2016).

\bibitem{shafiee2017evolution}
M.~J. Shafiee, E.~Barshan, and A.~Wong, ``Evolution in groups: A deeper look at
  synaptic cluster driven evolution of deep neural networks,'' {\em arXiv
  preprint arXiv:1704.02081}   (2017).

\bibitem{Armato1}
S.~G. Armato~III, G.~McLennan, L.~Bidaut, {\em et~al.}, ``The lung image
  database consortium (lidc) and image database resource initiative (idri): a
  completed reference database of lung nodules on ct scans,'' {\em Medical
  physics} {\bf 38}(2), 915--931  (2011).

\bibitem{Armato2}
S.~G. Armato~III, G.~McLennan, M.~F. McNitt-Gray, {\em et~al.}, ``Lung image
  database consortium: Developing a resource for the medical imaging research
  community 1,'' {\em Radiology} {\bf 232}(3), 739--748  (2004).

\bibitem{shafiee2016evolutionary}
M.~J. Shafiee and A.~Wong, ``Evolutionary synthesis of deep neural networks via
  synaptic cluster-driven genetic encoding,'' {\em arXiv preprint
  arXiv:1609.01360}   (2016).

\bibitem{lecun1998}
Y.~LeCun, L.~Bottou, Y.~Bengio, {\em et~al.}, ``Gradient-based learning applied
  to document recognition,'' {\em Proceedings of the IEEE}   (1998).

\bibitem{Ngiam2011}
J.~Ngiam, A.~Coates, A.~Lahiri, {\em et~al.}, ``On optimization methods for
  deep learning,'' in {\em Proceedings of the 28th International Conference on
  Machine Learning (ICML-11)},  265--272  (2011).

\bibitem{toyoda2008sensitivity}
Y.~Toyoda, T.~Nakayama, Y.~Kusunoki, {\em et~al.}, ``Sensitivity and
  specificity of lung cancer screening using chest low-dose computed
  tomography,'' {\em British journal of cancer} {\bf 98}(10), 1602--1607
  (2008).

\end{thebibliography}
\bibliographystyle{spiejour}   


\listoffigures
\listoftables

\end{spacing}
\end{document}